\documentclass[10pt,twocolumn,letterpaper]{article}
\usepackage{wacv}
\usepackage{times}
\usepackage{epsfig}
\usepackage{graphicx}
\usepackage{amsmath}
\usepackage{amssymb}
\usepackage{comment}
\usepackage{enumitem}
\usepackage{cite}
\usepackage[subrefformat=parens]{subcaption}
\usepackage{booktabs}
\usepackage{multirow}
\usepackage{arydshln}

\newcommand{\argmax}{\mathop{\rm arg~max}\limits}

\usepackage[pagebackref=true,breaklinks=true,letterpaper=true,colorlinks,bookmarks=false]{hyperref}

\wacvfinalcopy

\begin{document}

\title{Augmented Hard Example Mining for Generalizable Person Re-Identification}

\author{Masato Tamura \\
Hitachi, Ltd.
\and
Tomokazu Murakami \\
Hitachi, Ltd.
\and
{\tt\small \{masato.tamura.sf, tomokazu.murakami.xr\}@hitachi.com}
}

\maketitle

\begin{abstract}
Although the performance of person re-identification (Re-ID) has been much improved by using sophisticated training methods and large-scale labelled datasets, many existing methods make the impractical assumption that information of a target domain can be utilized during training. In practice, a Re-ID system often starts running as soon as it is deployed, hence training with data from a target domain is unrealistic. To make Re-ID systems more practical, methods have been proposed that achieve high performance without information of a target domain. However, they need cumbersome tuning for training and unusual operations for testing. In this paper, we propose augmented hard example mining, which can be easily integrated to a common Re-ID training process and can utilize sophisticated models without any network modification. The method discovers hard examples on the basis of classification probabilities, and to make the examples harder, various types of augmentation are applied to the examples. Among those examples, excessively augmented ones are eliminated by a classification based selection process. Extensive analysis shows that our method successfully selects effective examples and achieves state-of-the-art performance on publicly available benchmark datasets.
\end{abstract}

\section{Introduction}
Re-ID has received much attention thanks to its diverse applications such as surveillance and marketing. In Re-ID, pedestrian images across non-overlapping cameras are matched by features extracted from the images. Since the appearances of images drastically change due to variations in illumination, viewpoints, poses, and occlusions, it is difficult to acquire an identical feature from various images of the same pedestrian. To overcome this problem, many sophisticated methods have been proposed in the past few years~\cite{gprid_cvpr2019, gprid_bmvc2019, pd_uda_eccv2018, pd_uda_cvpr2018_1, pd_uda_bmvc2018, pd_cvpr2018, pd_uda_cvpr2018_2, s_aaai17, cuhk03_cvpr2014, s_aaai2018, uda_eccv2018, s_cvpr2016, person_search_cvpr2017, s_null_cvpr2016, s_ensembles_cvpr2015, cuhk02_cvpr2013, grid_cvpr2009, s_oneshot_cvpr2017, s_jlml_ijcai2017, s_imptrploss_cvpr2016, s_ssm_cvpr2017, market_iccv2015, s_spindle_cvpr2017, duke_iccv2017, viper_eccv2008, uda_uctl_cvpr2016}, and new approaches are being developed.

\begin{figure}[t]
    \centering
    \includegraphics[keepaspectratio, width=0.98\linewidth]{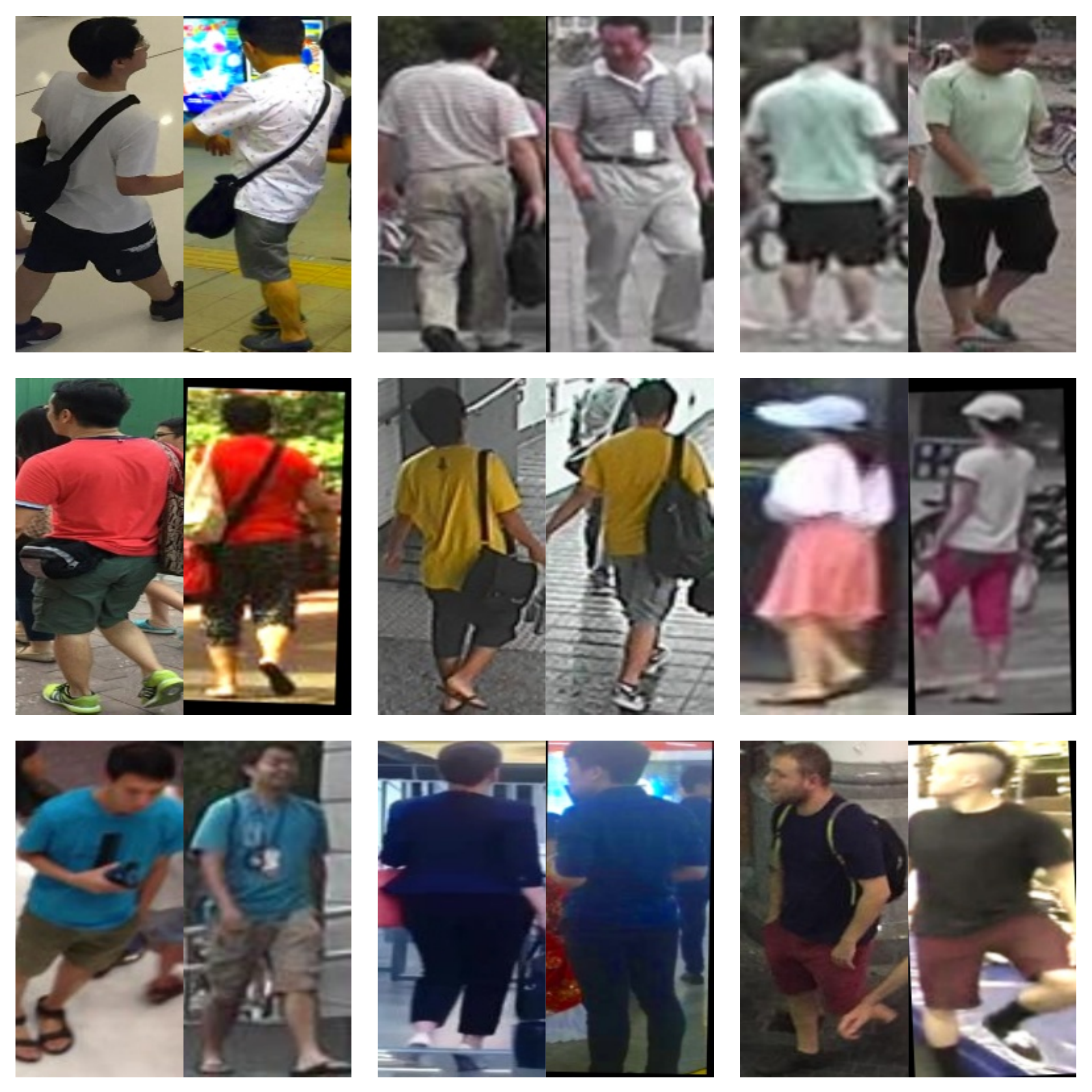}
    \caption{Examples of augmented hard examples. In each pair, the left image is the key of selecting hard examples, and the right image is a selected hard example to which various types of augmentation are applied.}
    \label{fig:example}
\end{figure}

Many existing approaches assume that data from a target domain are available during training. Some of these approaches undergo supervised training with data from a target domain~\cite{ s_aaai17, s_aaai2018, s_cvpr2016, s_null_cvpr2016, s_ensembles_cvpr2015, s_oneshot_cvpr2017, s_jlml_ijcai2017, s_imptrploss_cvpr2016, s_ssm_cvpr2017, s_spindle_cvpr2017}. They exhibit great performance when large-scale labelled datasets are prepared~\cite{cuhk02_cvpr2013,cuhk03_cvpr2014, person_search_cvpr2017, market_iccv2015, duke_iccv2017}, but these approaches suggest that a large number of annotations is needed on each deployment. Since the annotation process is time consuming, these supervised approaches are infeasible for practical use. To obviate the need for annotation on each deployment, unsupervised domain adaptation (UDA) approaches have recently been proposed~\cite{pd_uda_eccv2018, pd_uda_cvpr2018_1, pd_uda_bmvc2018, pd_uda_cvpr2018_2, uda_eccv2018, uda_uctl_cvpr2016}. These approaches adapt source domains to target domains by image translation, feature alignment, or multi-task learning. By this adaptation, domain-specific knowledge acquired from large-scale labelled datasets can be utilized for unlabelled datasets. The UDA approaches are more practical than the supervised approaches, but they still need data from a target domain during training.

In practice, data from a target domain are often unavailable until deployment, hence Re-ID models have to be trained only with data in existing domains and to match identities in an unseen domain. This setting is categorized as domain generalization (DG). If Re-ID models are simply trained in a supervised manner, the domain shift between training and testing is reported to substantially degrade performance~\cite{pd_uda_eccv2018, pd_uda_cvpr2018_1, pd_uda_bmvc2018, pd_cvpr2018, pd_uda_cvpr2018_2}, which suggests that the trained Re-ID models are over-fitted and have poor generalization performance. To solve this problem, a few methods have been proposed~\cite{ gprid_cvpr2019, gprid_bmvc2019}. They successfully improve Re-ID accuracy by adding some operations in MobileNetV2~\cite{mobilenetv2_cvpr2018}. However, modifying a sophisticated model requires cumbersome tuning for training until satisfactory performance can be achieved. Furthermore, additional operations slow inference speed, which is a significant drawback in practical applications. Considering this, sophisticated models should not be modified.

In this paper, we propose a data augmentation based method that can enhance the generalization performance without any network modification. The problem of data augmentation lies in determining augmentation policies. If the discrepancy between the statistics of augmented images and those of real images is huge, the performance will degrade. To solve this problem, automatic augmentation methods have been proposed~\cite{autoaug_cvpr2019, smart_aug_ieee, bayse_aug_nips2017}. These methods are learning based approaches. Effective augmentation policies can be learned by the methods. However, complex training procedures are needed in addition to original task procedures. Different from these methods, we adopt a simple selection strategy that does not need to learn about augmentation. In our method, first, hard examples are sampled on the basis of classification probabilities of an input mini-batch. Then, various types of augmentation are applied to make the examples harder. Finally, the hardest example is selected from them. Input mini-batches are augmented with only random horizontal flipping, hence our model is basically trained with realistic images. Thanks to this, excessively augmented examples are eliminated in the final selection process. Figure~\ref{fig:example} shows that our method successfully selects appropriate augmented examples.

Finally, we summarize our contributions as follows:
\begin{itemize}
\setlength{\parskip}{0cm}
\setlength{\itemsep}{0cm}
\item We propose a simple selection strategy for data augmentation, which eliminates excessively augmented images. Since our method needs only one ordinary network for training and testing, existing highly optimized models can be utilized without any network modification.
\item We investigate not only model accuracy but also computation cost for practical use.
\item We demonstrate state-of-the-art performance on Re-ID benchmarks and the robustness of our method against changes of augmentation parameters.
\end{itemize}

\section{Related work}
\subsection{Domain generalized person re-identification}
Although Re-ID has been researched for years, only a few methods focus on generalization performance~\cite{gprid_cvpr2019, gprid_bmvc2019}. In \cite{gprid_cvpr2019}, Song et al.\  proposed a meta-learning~\cite{meta} based model called domain-invariant mapping network (DIMN). Different from a common way that uses feature distances for matching scores, DIMN generates classifier weights from gallery images and then takes the dot product of the classifier weights and probe image features to calculate matching scores. This meta-learning pipeline makes the model domain-invariant, but the complicated meta-learning procedures make optimization difficult. In addition, classifier weight generation during testing slows the inference speed. Considering these drawbacks, a simpler approach that utilizes normalization was proposed by Jia et al.~\cite{gprid_bmvc2019}. They regard style and content variations as the cause of domain bias and suppress them by inserting instance normalization (IN)~\cite{instance_norm} to bottlenecks in shallow layers and a batch normalization (BN)~\cite{batch_norm} to a feature extraction layer. The evaluation results show that normalization successfully eliminates domain bias and improves the accuracy, but they also show that the location and amount of IN are important. As stated by Nam and Kim~\cite{batch_instance_norm}, insertion of instance normalization should be carefully investigated because excessive normalization suppresses styles that are the key factors to discriminate objects. This investigation process is cumbersome. Furthermore, both IN and BN add computation cost, hence inference speed slows down. Different from these methods, our method adopts a data augmentation based method and does not need any network modification, which makes our method more practical.

\subsection{Automatic data augmentation}
\begin{figure*}[t]
    \centering
    \includegraphics[keepaspectratio, width=1.0\linewidth]{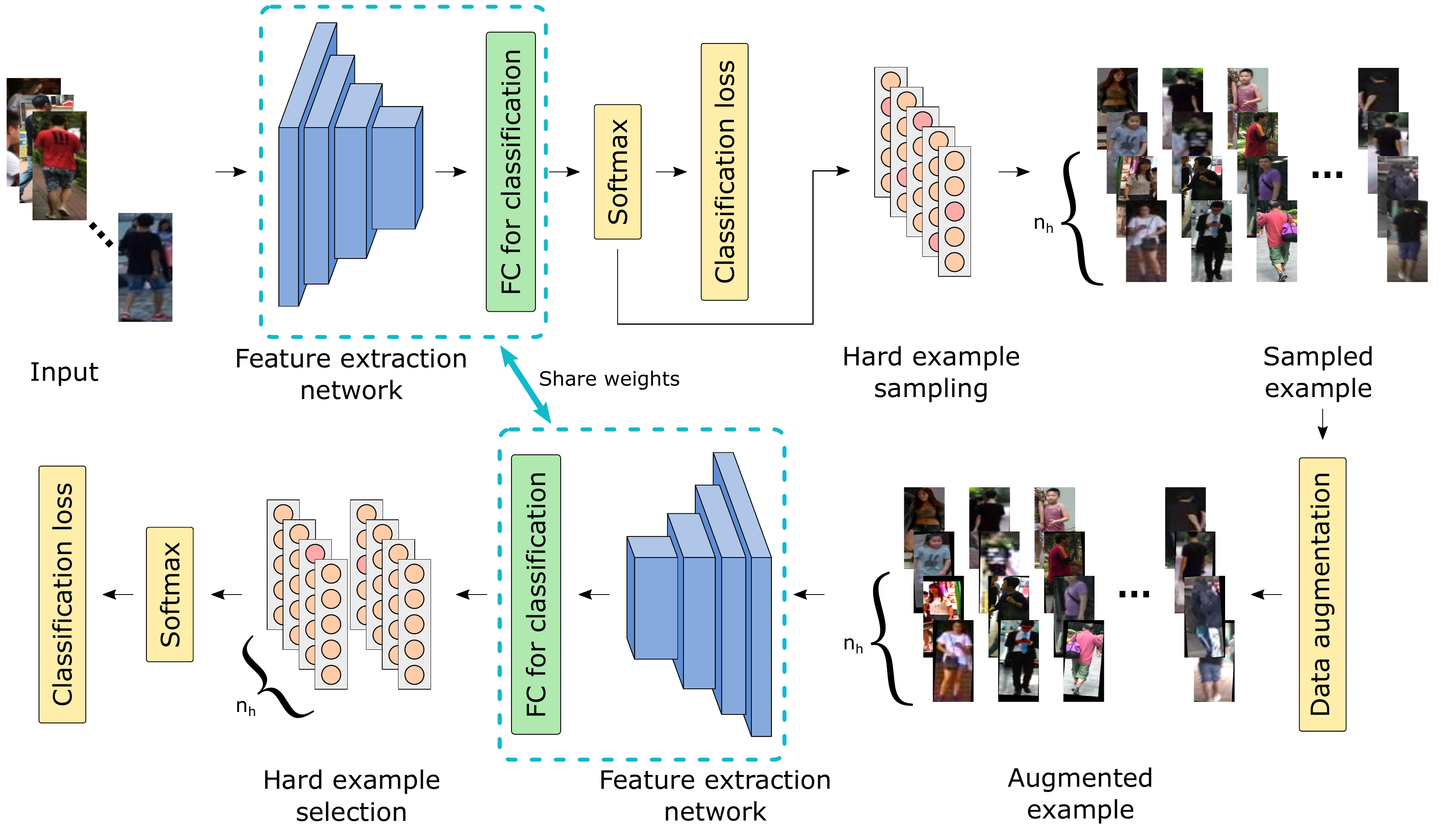}
    \caption{Structure of the proposed method.}
    \label{fig:structure}
\end{figure*}

Although data augmentation effectively enhances generalization performance, the types and their parameters are difficult to determine. In common cases, they are determined by intuition or trial-and-error operation with validation images, but the results of this approach are unstable and troublesome. To solve this problem, automatic data augmentation methods have been proposed~\cite{autoaug_cvpr2019, smart_aug_ieee, bayse_aug_nips2017}. Lemley et al.~\cite{smart_aug_ieee} proposed a network that merges two or more samples to generate an augmented sample. Tran et al.~\cite{bayse_aug_nips2017} also proposed a generation based method, but they used a Bayesian approach and generative adversarial networks~\cite{gan} for generation. Different from these two methods, Cubuk et al.~\cite{ autoaug_cvpr2019} employed a searching strategy. In this method, appropriate data augmentation policies are investigated by reinforcement learning~\cite{rl} with a recurrent neural network controller. All the methods improve performance, but they need additional networks that have to be trained for data augmentation. This makes original task training complicated. Unlike these methods, our method does not need any additional networks for data augmentation and can be easily integrated into a common training process.

\section{Proposed method}
\subsection{Overview}
For setting domain generalized Re-ID, we assume that we have $K$ source domains (datasets) $\mathcal{D} = \left\{D_i \mid i = 1, 2, ..., K\right\}$. Each domain $D_i = \left\{X^{\left(i\right)}, Y^{\left(i\right)}\right\}$ contains image-label pairs and has its own label space $y^{\left(i\right)} \in \left\{l_{j}^{\left(i\right)} \mid j = 1, 2, ..., {M^{\left(i\right)}}\right\}$, where $M^{\left(i\right)}$ is the number of identities in $D_i$. Since each label space is disjointed from others, we take the union of the label spaces for a training label space. As a result, the size of the label space becomes as follows:
\begin{equation}
    N = \sum_{i=1}^{K} M^{\left(i\right)}.
\end{equation}

For a simple yet strong baseline, we take a naive deep learning approach called aggregation (AGG). In AGG, a model is trained to minimize cross-entropy (CE) loss of all identities from all domains:
\begin{equation}\label{eq:ce}
 L_{CE} = \frac{1}{n_{bs}}\sum_{i=1}^{n_{bs}} l^{\left(CE\right)}\left(g_{\phi}\left(f_{\theta}\left(x_i\right)\right), y_i\right).
\end{equation}
Here, $n_{bs}$ is the number of images in a mini-batch, $f_{\theta}$ is a feature extractor parameterized by $\theta$, and $g_{\phi}$ is a classifier parameterized by $\phi$. After training, the feature extractor $f_{\theta}$ is used to extract features from images. Then the features are used to calculate matching scores as follows:
\begin{equation}
 s = 1 - \frac{\lVert \hat{\boldsymbol{z}}_{\boldsymbol{p}} - \hat{\boldsymbol{z}}_{\boldsymbol{g}} \rVert}{2}.
\end{equation}
Here, $\hat{\boldsymbol{z}}_{\boldsymbol{p}}$ is an L2 normalized feature of a probe image, and $\hat{\boldsymbol{z}}_{\boldsymbol{g}}$ is that of a gallery image. Gallery images having high scores are considered to be the images of the same identity in the probe image.

On the basis of this baseline, we propose two methods: hard example mining with CE loss (Sec.\ref{subsec:hard_mining}) and augmented hard example selection (Sec.\ref{subsec:aug_selection}). We explain them in the following sections.

\subsection{Hard example mining with cross entropy loss}\label{subsec:hard_mining}
Hard example mining is a way to improve the performance in borderline cases and enhance generalization performance. Although the importance of hard example mining with triplet loss is mentioned by Hermans et al.~\cite{triplet}, CE loss based hard example mining has never been explored. In this section, we explain how to select hard examples during training with CE loss.

As shown in Fig.~\ref{fig:structure}, first, a mini-batch is input to a feature extraction network, and then the extracted features are input to a fully-connected (FC) layer. Classification loss of the mini-batch $L_{batch}$ is calculated in accordance with Eq.~\ref{eq:ce}. The outputs of Softmax are recalculated for hard example mining. The recalculation is as follows:
\begin{equation}\label{eq:prob}
 \Pr\left(i \mid x_{id=j}\right) = \frac{\mathrm{e}^{p_i}}{\sum_{k\neq j} \mathrm{e}^{p_k}} \hspace{5mm} (i \neq j).
\end{equation}
Here, $i$ is the class index of an identity, $j$ is the class index of an identity in the input mini-batch, and $p$ is the output of the FC layer. On the basis of this probability, hard examples against identities in the input mini-batch are selected. Concretely, $n_h$ identities are sampled with replacement in accordance with the probability calculated by Eq.~\ref{eq:prob}, and then an image of each sampled identity is randomly selected from the images of the identity. A new mini-batch is created by collecting images of hard examples sampled against all the identities in the input mini-batch, and CE loss of the new mini-batch is calculated in the same iteration.

In the initial stage of training, there are no clues for Re-ID, hence the described sampling method works as random sampling. As the training proceeds, the probability calculated by Eq.~\ref{eq:prob} indicates the similarity of identities, hence the method works as hard example mining. These characteristics well fit deep learning training.

\subsection{Augmented hard example selection}\label{subsec:aug_selection}
Our method has two augmentation policies: one for input mini-batches, and the other for hard examples. Since random horizontal flipping exactly has a positive impact, it is included in both policies. As for other augmentation methods, the impacts are unknown, hence they are applied to only hard examples, and excessively augmented examples are eliminated in a selection process.

As shown in Fig.~\ref{fig:structure}, images of a mini-batch created by the hard example mining are augmented before being input to a feature extraction network. Then, the augmented images are input to the network and classified by the FC layer. Note that the weights of the network and the FC layer are shared with those of the network and the FC layer used for an input mini-batch. For selecting appropriate hard examples, the outputs of the FC layer are used.

Suppose that there are $n_h$ images of identities sampled as hard examples against one identity whose class index is $i$, and outputs of the images from the FC layer are denoted by $\mathcal{Q}=\left\{\boldsymbol{q^{\left(i,j\right)}} \mid j = 1, 2, ..., n_h \right\}$. From the outputs, one output is selected as follows:
\begin{equation}
 k_{i} = \argmax_{j} q_{i}^{\left(i,j\right)}.
\end{equation}
Here, $q^{\left(i,j\right)}_i$ is the $i$-th entry of $\boldsymbol{q^{\left(i,j\right)}}$. By using the selected outputs, loss of the augmented hard examples is calculated as follows:
\begin{equation}
 L_{aug} = \frac{1}{n_{bs}}\sum_{i=1}^{n_{bs}} l^{\left(CE\right)}\left(\boldsymbol{q^{\left(i,k_{i}\right)}}, y^{\left(i,k_{i}\right)}\right).
\end{equation}
Here, $n_{bs}$ is the number of images in a mini-batch, and $y^{\left(i,k_{i}\right)}$ is the label of the selected hard example.

Since an example is selected from augmented hard examples on the basis of the similarity of identities, the example can be harder, and at the same time, excessively augmented examples are eliminated. By combining hard example mining and data augmentation, the two methods work complementarily. As a result, a trained model can robustly discriminate similar identities.

Finally, total loss is calculated as follows:
\begin{equation}
    L_{total} = \frac{\left(L_{batch} + L_{aug}\right)}{2}.
\end{equation}

\section{Experiments}
\subsection{Datasets and evaluation settings}\label{subsec:dataset}
\begin{table}[t]
  \caption{Dataset statistics.}
  \begin{subtable}{1.0\linewidth}
    \subcaption{Training datasets.}\label{table:train_dataset}
    \centering
    \begin{tabular}{lcc}
      \toprule
      Dataset & \#IDs & \#Images  \\
      \midrule
      CUHK02 & 1,816 & 7,264 \\
      CUHK03 & 1,467 & 14,097 \\
      Duke MTMC & 1,812 & 36,411 \\
      Market1501 & 1,501 & 29,419 \\
      PersonSearch & 11,934 & 34,574 \\
      \midrule
      & 18,530 & 121,765 \\
      \bottomrule
    \end{tabular}
    \medskip
  \end{subtable}
  \begin{subtable}{1.0\linewidth}
    \subcaption{Test datasets. (``Pr.": Probe, ``Ga".: Gallery)}
    \label{table:test_dataset}
    \centering
    \begin{tabular}{lcccc}
      \toprule
      Dataset & \#Pr. IDs & \#Ga. IDs & \#Pr. images & \#Ga. images \\
      \midrule
      VIPeR & 316 & 316 & 316 & 316 \\
      PRID & 100 & 649 & 100 & 649 \\
      GRID & 125 & 900 & 125 & 900 \\
      i-LIDS & 60 & 60 & 60 & 60 \\
      \bottomrule
    \end{tabular}
  \end{subtable}
\end{table}

To evaluate our method, we follow the settings described by Jia et al.~\cite{gprid_cvpr2019} and Song et al.~\cite{gprid_bmvc2019}. In the settings, large-scale datasets are combined to train Re-ID models, and small-scale datasets are individually used to evaluate model performance. The statistics of training and evaluation datasets are shown in Tables~\ref{table:train_dataset} and ~\ref{table:test_dataset}, respectively. For training, CUHK02~\cite{cuhk02_cvpr2013}, CUHK03~\cite{cuhk03_cvpr2014}, Duke MTMC~\cite{duke_iccv2017}, Market1501~\cite{market_iccv2015}, and PersonSearch~\cite{person_search_cvpr2017} are used. All the datasets have more than a thousand identities and thousands of images. By combining the datasets, Re-ID models are trained with 121,765 images of 18,530 identities. For evaluation, VIPeR~\cite{viper_eccv2008}, PRID~\cite{prid_scia2011}, GRID~\cite{grid_cvpr2009}, and i-LIDS~\cite{ilids_bmvc2009} are used. They are relatively small-scale datasets and have at most a thousand identities. From the identities in each dataset, probe identities and gallery identities are randomly sampled in accordance with the number shown in Table~\ref{table:test_dataset}. With the sampled identities, Re-ID models are evaluated in a single-shot Re-ID manner. We do the sampling and the evaluation 10 times for each dataset and average the results.

\subsection{Evaluation metrics}
To show Re-ID model performance, we use cumulative matching characteristics (CMC).
CMC shows Re-ID accuracy for each rank $k$. $k$ is set to 1, 5, and 10.

\subsection{Implementation details}
\begin{table}[t]
  \caption{The types of data augmentation applied to images sampled in hard example mining.}
  \label{table:data_aug}
  \centering
  \begin{tabular}{llr}
    \toprule
    Type & Parameter &\\
    \midrule
    Random crop & Edge offset: & $-10$--$10$\\
    \hdashline[2.0pt/2.0pt] \\[-1.8\medskipamount]
    Random horizontal flip & Probability: & 0.5\\
    \hdashline[2.0pt/2.0pt] \\[-1.8\medskipamount]
    Random rotation & Degree: & $-5^\circ$--$5^\circ$\\
    \hdashline[2.0pt/2.0pt] \\[-1.8\medskipamount]
    \multirow{3}{*}{Random color jitter} & Hue value: & $-0.1$--$0.1$\\
    & Saturation scale: & 0.5--2.0 \\
    & Value scale: & 0.5--2.0\\
    \bottomrule
  \end{tabular}
\end{table}

We use MobileNetV2~\cite{mobilenetv2_cvpr2018} as a feature extraction network. Two width multipliers, which are 0.75 and 1.0, are used to analyze computation cost. For training, the FC layer of the original MobileNetV2 is replaced with a FC layer that has units equal in number to the identities in the training datasets. The network is fine-tuned from weights pretrained on ImageNet~\cite{imagenet_cvpr2009} using the combined dataset described in Sec.~\ref{subsec:dataset} for 30 epochs. The initial learning rate is set to 0.01 and decayed by 0.1 after 20 epochs. To optimize the network, we use stochastic gradient descent with momentum, which is set to 0.9. Input images are resized to $256 \times 128$. Batch size is set to 16. To prevent over-fitting, weight decay, label smoothing~\cite{duke_iccv2017, smooth_cvpr2019}, dropout~\cite{dropout_jmlr}, and data augmentation are used. The weight decay rate is set to 0.0005, smoothing value is set to 0.1, and dropout rate is set to 0.5. As for data augmentation, images in input mini-batches are horizontally flipped with a probability of 0.5. On the other hand, images selected in hard example mining are augmented by various types of augmentation, which are detailed in Table~\ref{table:data_aug}. Random cropping, random flipping, random rotation, and random color jitter are used. If a parameter is denoted with a range, an applied value is uniformly sampled within the range at every augmentation process. For random cropping, an offset from each edge is determined by a sampled value. If offset positions are the outside of an image, the image is padded with zero values. The number of augmented hard examples for each identity, which is denoted by $n_h$, is set to 4. To prevent all the augmented hard examples from being excessively augmented, one is augmented by only random horizontal flipping. For evaluation, extracted features are L2 normalized before matching scores are calculated. Note that we do not use any test-time data augmentation.

\subsection{Comparison against state-of-the-art}
\begin{table*}[t]
  \caption{Comparison results against baselines. (``R": Rank, ``S": Supervised training with a target dataset, ``U": UDA, ``DG": Domain generalization, ``-": No report)}
  \label{table:comp_sota}
  \centering
  \begin{tabular}{lccccccccccccc}
    \toprule
    &&\multicolumn{3}{c}{VIPeR}&\multicolumn{3}{c}{PRID}&\multicolumn{3}{c}{GRID}&\multicolumn{3}{c}{i-LIDS}\\
    \cmidrule(lr){3-5}\cmidrule(lr){6-8}\cmidrule(lr){9-11}\cmidrule(lr){12-14}
    Method & Type & R-1 & R-5 & R-10 & R-1 & R-5 & R-10 & R-1 & R-5 & R-10 & R-1 & R-5 & R-10 \\
    \midrule
    Ensembles~\cite{s_ensembles_cvpr2015} & S & 45.9 & 77.5 & 88.9 & 17.9 & 40.0 & 50.0 & - & - & - & 50.3 & 72.0 & 82.5 \\
    DNS~\cite{s_null_cvpr2016} & S & 42.3 & 71.5 & 82.9 & 29.8 & 52.9 & 66.0 & - & - & - & - & - & - \\
    ImpTrpLoss~\cite{s_imptrploss_cvpr2016} & S & 47.8 & 74.4 & 84.8 & 22.0 & - & 47.0 & - & - & - & 60.4 & 82.7 & 90.7 \\
    GOG~\cite{s_cvpr2016} & S & 49.7 & \textbf{79.7} & 88.7 & - & - & - & 24.7 & 47.0 & 58.4 & - & - & - \\
    MTDnet~\cite{s_aaai17} & S & 47.5 & 73.1 & 82.6 & 32.0 & 51.0 & 62.0 & - & - & - & 58.4 & 80.4 & 87.3 \\
    OneShot~\cite{s_oneshot_cvpr2017} & S & 34.3 & - & - & 41.4 & - & - & - & - & - & 51.2 & - & - \\
    SpindleNet~\cite{s_spindle_cvpr2017} & S & 53.8 & 74.1 & 83.2 & \textbf{67.0} & \textbf{89.0} & \textbf{89.0} & - & - & - & 66.3 & 86.6 & 91.8 \\
    SSM~\cite{s_ssm_cvpr2017} & S & 53.7 & - & \textbf{91.5} & - & - & - & 27.2 & - & 61.2 & - & - & - \\
    JLML~\cite{s_jlml_ijcai2017} & S & 50.2 & 74.2 & 84.3 & - & - & - & 37.5 & 61.4 & 69.4 & - & - & - \\
    \midrule
    UCTL~\cite{uda_uctl_cvpr2016} & U & 31.5 & - & - & 24.2 & - & - & - & - & - & 49.3 & - & - \\
    TJAIDL~\cite{pd_uda_cvpr2018_2} & U & 38.5 & - & - & 34.8 & - & - & - & - & - & - & - & - \\
    MMFAN~\cite{pd_uda_bmvc2018} & U & 39.1 & - & - & 35.1 & - & - & - & - & - & - & - & - \\
    Synthesis~\cite{uda_eccv2018} & U & 43.0 & - & - & 43.0 & - & - & - & - & - & 56.5 & - & - \\
    \midrule
    AGG (DIMN)~\cite{gprid_cvpr2019} & DG & 42.9 & 61.3 & 68.9 & 38.9 & 63.5 & 75.0 & 29.7 & 51.1 & 60.2 & 69.2 & 84.2 & 88.8 \\
    AGG (DualNorm)~\cite{gprid_bmvc2019} & DG & 42.1 & - & - & 27.2 & - & - & 28.6 & - & - & 66.3 & - & - \\
    AGG (Ours) & DG & 42.4 & 61.1 & 69.2 & 22.3 & 45.2 & 54.3 & 31.4 & 49.8 & 58.7 & 69.8 & 88.3 & 93.5 \\
    DIMN~\cite{gprid_cvpr2019} & DG & 51.2 & 70.2 & 76.0 & 39.2 & 67.0 & 76.7 & 29.3 & 53.3 & 65.8 & 70.2 & 89.7 & 94.5 \\
    DualNorm~\cite{gprid_bmvc2019} & DG & \textbf{53.9} & - & - & 60.4 & - & - & 41.4 & - & - & 74.8 & - & - \\
    Ours & DG & 49.8 & 70.8 & 77.0 & 34.3 & 56.2 & 65.7 & \textbf{46.6} & \textbf{67.5} & \textbf{76.1} & \textbf{76.3} & \textbf{93.0} & \textbf{95.3} \\
    \bottomrule
  \end{tabular}
\end{table*}
\begin{table*}[t]
  \caption{Computation cost comparison against DG baselines. (``W": Width multiplier, ``MAdd": Multiply-adds, ``R": Rank)}
  \label{table:comp_cost}
  \centering
  \begin{tabular}{lcrccccccccccccc}
    \toprule
    &&&&\multicolumn{3}{c}{VIPeR}&\multicolumn{3}{c}{PRID}&\multicolumn{3}{c}{GRID}&\multicolumn{3}{c}{i-LIDS}\\
    \cmidrule(lr){5-7}\cmidrule(lr){8-10}\cmidrule(lr){11-13}\cmidrule(lr){14-16}
    Method & W & MAdd & Time & R-1 & R-5 & R-10 & R-1 & R-5 & R-10 & R-1 & R-5 & R-10 & R-1 & R-5 & R-10 \\
    \midrule
    DIMN~\cite{gprid_cvpr2019} & 1.4 & 1523M & 2.23 ms & 51.2 & 70.2 & 76.0 & 39.2 & \textbf{67.0} & \textbf{76.7} & 29.3 & 53.3 & 65.8 & 70.2 & 89.7 & 94.5 \\
    DualNorm~\cite{gprid_bmvc2019} & 1.0 & 791M & 2.68 ms & \textbf{53.9} & - & - & \textbf{60.4} & - & - & 41.4 & - & - & 74.8 & - & - \\
    Ours & 0.75 & \textbf{543M} & \textbf{2.06 ms} & 49.6 & 69.6 & 75.2 & 33.5 & 51.7 & 63.0 & 41.1 & 61.3 & 69.0 & \textbf{77.2} & 91.3 & 95.0 \\
    Ours & 1.0 & 783M & 2.10 ms & 49.8 & \textbf{70.8} & \textbf{77.0} & 34.3 & 56.2 & 65.7 & \textbf{46.6} & \textbf{67.5} & \textbf{76.1} & 76.3 & \textbf{93.0} & \textbf{95.3} \\
    \bottomrule
  \end{tabular}
\end{table*}

To demonstrate the superiority of our method, we compare it with previously proposed baselines.
For Re-ID, three types of approaches have been proposed. The types are as follows:
\begin{description}[style=unboxed,leftmargin=0cm,noitemsep,topsep=0pt]
\item[Supervised training with a target dataset] This is the most basic type and has been researched for years. Although high performance is realized with a large-scale dataset, the performance is still low with a small-scale dataset. To solve this problem, many methods have been proposed~\cite{s_ensembles_cvpr2015, s_null_cvpr2016, s_imptrploss_cvpr2016, s_cvpr2016, s_aaai17, s_oneshot_cvpr2017, s_spindle_cvpr2017, s_ssm_cvpr2017, s_jlml_ijcai2017}. The upper part of Table~\ref{table:comp_sota} shows their benchmark results. Among them, SpindleNet~\cite{s_spindle_cvpr2017}, SSM~\cite{s_ssm_cvpr2017}, and JLML~\cite{s_jlml_ijcai2017} perform well. Since they have different settings from our method, fair comparison is difficult. However, we show their results as references. Except for PRID, our method shows competitive or even better results. This means that domain specific characteristics can be covered by combining multiple large-scale datasets and appropriate data augmentation.
\item[Unsupervised domain adaptation] The purpose of UDA is to transfer knowledge from large-scale labelled datasets to unlabelled datasets. In accordance with this purpose, some UDA approaches have been proposed for Re-ID~\cite{pd_uda_bmvc2018, pd_uda_cvpr2018_2, uda_eccv2018, uda_uctl_cvpr2016}. The middle part of Table~\ref{table:comp_sota} shows their benchmark results. Synthesis~\cite{uda_eccv2018} performs the best among them by utilizing a synthetic dataset. The same as the supervised training with a target dataset, the UDA methods have different settings from ours. However, we show their results as references. For all the benchmark datasets, our method outperforms the UDA methods. This means that our method can competitively utilize large-scale datasets.
\item[Domain generalization] DG setting has the most practical assumption that a target dataset cannot be seen during training. Because of this setting, DG methods have to learn general feature representation from existing datasets. For this purpose, a few methods have been proposed~\cite{gprid_cvpr2019, gprid_bmvc2019}, and our method is also evaluated under this setting. The lower part of Table~\ref{table:comp_sota} shows the benchmark results of the methods. In this comparison, we set the width multiplier of MobileNetV2 to 1.0. We put the AGG result of each method to show that the baselines are almost the same in all the methods. DualNorm~\cite{gprid_bmvc2019} outperforms the others for VIPeR and PRID, while our method outperforms the others for GRID and i-LIDS. These results demonstrate the effectiveness of our method.
\end{description}

\subsection{Computation cost analysis}
For practical use, inference time is an important factor for Re-ID performance. We compare the computation cost and the inference time of the models in the DG setting. For fair comparison, we set the input image size to $256 \times 128$ for all the methods. Table~\ref{table:comp_cost} shows the comparison results. Multiply-adds (MAdd) is estimated by Tensorflow profiler~\cite{tensorflow}. For calculating inference time, we use RTX 2080Ti with CUDA ver. 10.0~\cite{cuda}. The inference time is the time it takes to calculate a matching score for one pair of a probe image and a gallery image. To analyze computation cost, the results of our method with 0.75 and 1.0 width multipliers are shown in the table.

Our method has a shorter inference time than the other two, because we do not add any operations to the original MobileNetV2. The magnitude of the difference is only 0.1 ms, but it accumulates while matching scores are calculated for all the pairs of probe images and gallery images. Considering this, our method is more practical than the other two.

The difference in inference time between DuanlNorm and ours with the 1.0 width multiplier (2.68 vs. 2.10 ms) is larger than that between ours with the 0.75 and 1.0 width multipliers (2.06 vs. 2.10 ms) even though MAdd of DualNorm is almost the same as that of ours with the 1.0 width multiplier (791M vs. 783M). This means that the instance normalization causes slow inference speed. In general, unusual operations are not optimized for high speed computation in usual deep learning libraries, hence they take a long time regardless of computation cost. Since the optimization process is cumbersome, practical models should be composed of usual operations (e.g., convolution and batch normalization). From this point of view, our method has an advantage.

\subsection{Ablation study}
\begin{table}[t]
  \caption{Ablation study on the impact of different components. In the table, only rank-1 accuracy is shown.
  (``Aug.": Augmented)}
  \label{table:comp_abl}
  \centering
  \begin{tabular}{lcccc}
    \toprule
    Component & VIPeR & PRID & GRID & i-LIDS \\
    \midrule
    Baseline & 42.4 & 22.3 & 31.4 & 69.8 \\
    Augment & 43.0 & 29.0 & 36.4 & 71.2 \\
    Mining & 47.3 & 27.4 & 38.2 & 73.5 \\
    Augment + mining & 47.3 & 28.7 & 41.4 & 74.5 \\
    Aug. mining select & 49.8 & 34.3 & 46.6 & 76.3 \\
    \bottomrule
  \end{tabular}
\end{table}

To analyze the effect of each component in our method, we evaluate rank-1 accuracy with each component. In this evaluation, the width multiplier is set to 1.0. Table~\ref{table:comp_abl} shows the evaluation results. Each component is as follows:
\begin{description}[style=unboxed,leftmargin=0cm,noitemsep,topsep=0pt]
\item[Baseline] AGG.
\item[Augment] The proposed hard example mining is not carried out. Instead, all the images in input mini-batches are augmented by the methods shown in Table~\ref{table:data_aug}.
\item[Mining] The proposed hard example mining is carried out, but sampled hard examples are augmented by only random horizontal flipping. In this case, $n_h$ is set to 1, and the selection process is skipped.
\item[Augment + mining] Combination of \textbf{Augment} and \textbf{Mining}. This means that all the images input to a network are augmented by the methods shown in Table~\ref{table:data_aug}.
\item[Aug. mining select] The proposed method.
\end{description}

We can see from the results of \textbf{Augment} and \textbf{Mining} that both the data augmentation and the proposed hard example mining improve the generalization performance. However, the results of \textbf{Augment + mining} show that just combining the two methods does not improve the performance much from each method. Compared to \textbf{Augment + mining}, \textbf{Aug. mining select} has better effect on the performance. This means that our method successfully selects hard examples that have positive impact on the performance, and the proposed selection is important for the improvement.

In total, our method improves the rank-1 accuracy of \textbf{Baseline} by 7.4\%, 12.0\%, 15.2\%, and 6.5\% for VIPeR, PRID, GRID, and i-LIDS, respectively.

\subsection{Robustness of selection strategy}
\begin{table}[t]
  \caption{Data augmentation patterns. (``H": Hue value, ``S": Saturation scale, ``V": Value scale)}
  \label{table:data_aug_pattern}
  \centering
  \begin{tabular}{lcccc}
    \toprule
    && Weak & Moderate & Strong \\
    \midrule
    Crop && -5--5 & -10--10 & -15--15 \\
    \hdashline[2.0pt/2.0pt] \\[-1.8\medskipamount]
    Rotation &&$0^\circ$ &$-5^\circ$--$5^\circ$ &$-10^\circ$--$10^\circ$ \\
    \hdashline[2.0pt/2.0pt] \\[-1.8\medskipamount]
    \multirow{3}{*}{Color} & H & $-0.05$--$0.05$ & $-0.1$--$0.1$ & $-0.15$--$0.15$ \\
    & S & 0.67--1.5 & 0.5--2.0 & 0.4--2.5 \\
    & V & 0.67--1.5 & 0.5--2.0 & 0.4--2.5 \\
    \bottomrule
  \end{tabular}
\end{table}

\begin{table}[t]
  \caption{Comparison of three augmentation patterns with and without the proposed method. In the table, only rank-1 accuracy is shown. (``min.": mining)}
  \label{table:comp_aug}
  \centering
  \begin{tabular}{lcccc}
    \toprule
    Pattern & VIPeR & PRID & GRID & i-LIDS \\
    \midrule
    Weak w/o min. select & 45.2 & 30.2 & 37.2 & 72.0 \\
    Moderate w/o min. select & 43.0 & 29.0 & 36.4 & 71.2 \\
    Strong w/o min. select & 40.8 & 28.1 & 37.3 & 69.7 \\
    Weak w/ min. select & 49.2 & 36.2 & 43.0 & 76.2 \\
    Moderate w/ min. select & 49.8 & 34.3 & 46.6 & 76.3 \\
    Strong w/ min. select & 48.8 & 36.2 & 45.4 & 75.7 \\
    \bottomrule
  \end{tabular}
\end{table}

\begin{figure}[t]
    \centering
    \includegraphics[keepaspectratio, width=1.0\linewidth]{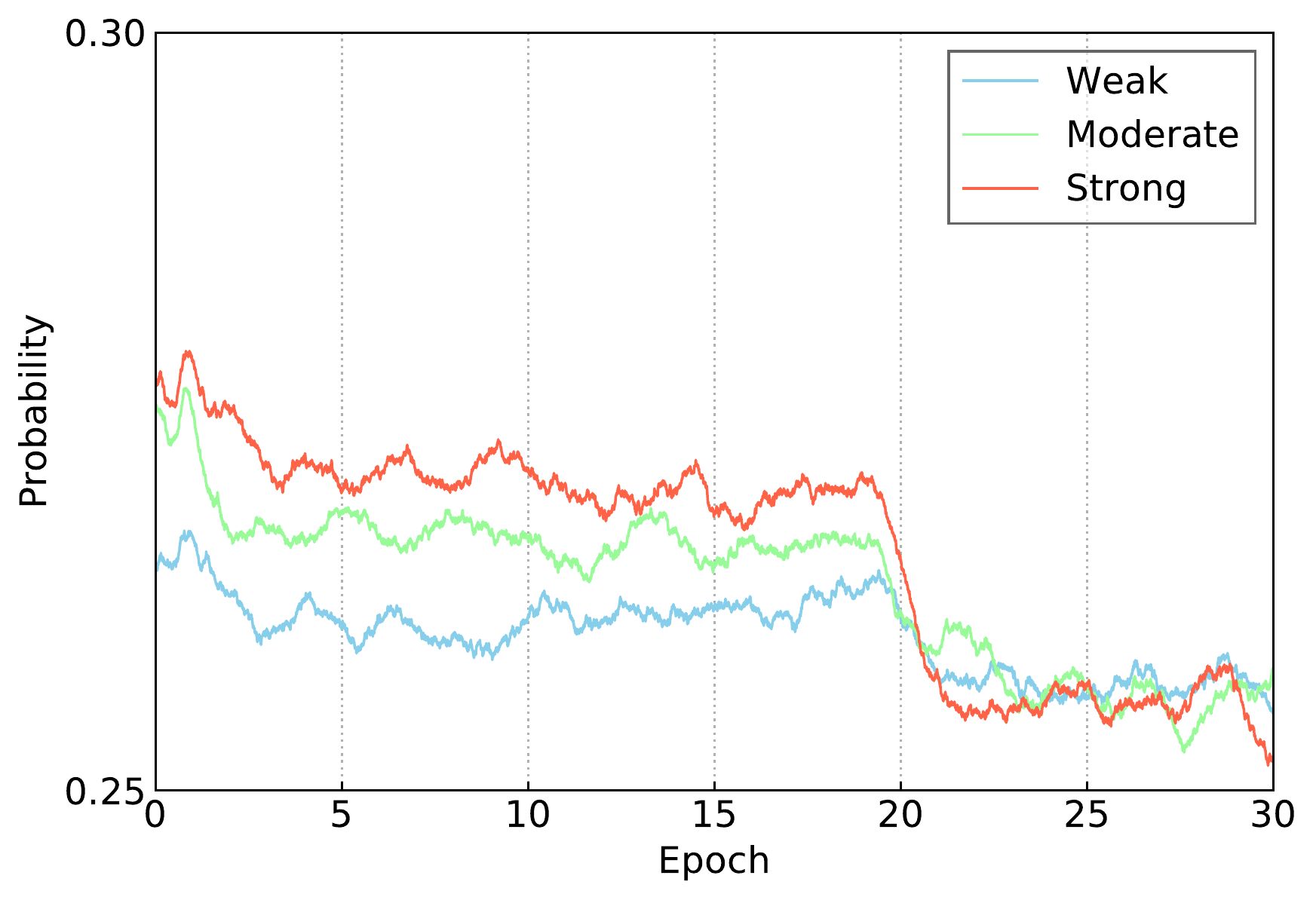}
    \caption{Probabilities of selecting an example augmented by only random horizontal flipping for three augmentation patterns.}
    \label{fig:prob}
\end{figure}

To show the robustness of our selection strategy, we train the model with three patterns of data augmentation and evaluate its performance. The patterns are shown in Table~\ref{table:data_aug_pattern}. The moderate data augmentation is the same as the augmentation shown in Table~\ref{table:data_aug}. The parameter ranges of weak and strong data augmentation are narrower and broader than that of moderate data augmentation, respectively. In all the patterns, random horizontal flopping is used with probability of 0.5.

Table~\ref{table:comp_aug} shows the evaluation results. The upper three rows show the results of usual input data augmentation with the three patterns, and the lower three rows show those of the proposed method with the three patterns. With the input data augmentation, stronger data augmentation degrades the performance except for GRID, whereas with the hard example data augmentation, stronger data augmentation does not change the performance much or even improves it. This shows that our method broadened the acceptable range of data augmentation and that our selection strategy is robust.

To further clarify this consideration, we examine the probability of selecting the example augmented by only random horizontal flipping. Figure~\ref{fig:prob} shows the probability. Since $n_h$ is set to 4, one example is supposed to be selected with the probability of 0.25, but the probability is higher. In addition, the stronger augmentation makes the probability higher until the learning rate is decayed. This means that as the probability of containing excessively augmented images becomes higher, the probability of selecting realistic images becomes higher. We can see from this result that our method works as intended. After the learning rate is decayed, the probabilities of all the augmentation patterns become the same. We will investigate the cause and effect of this in our future work.

\section{Conclusion}
In this paper, we have proposed a simple selection strategy for data augmentation to improve Re-ID performance. In our method, various augmentation methods are applied to only hard examples sampled on the basis of classification probabilities, and excessively augmented examples are eliminated as easy examples. Since our method uses classification probability for selection, it can be easily integrated into a common training process. In addition, our method does not need any unusual operations in networks, so highly optimized models can be utilized without any modification. Experiments on four public benchmark datasets show that our method can achieve state-of-the-art performance for practical use in Re-ID.

{\small
\bibliographystyle{ieee}
\bibliography{egpaper_final}
}

\end{document}